\documentclass[letterpaper]{article} 
\usepackage{aaai24}  
\usepackage{times}  
\usepackage{helvet}  
\usepackage{courier}  
\usepackage[hyphens]{url}  
\usepackage{graphicx} 
\usepackage{natbib}  
\usepackage{caption} 
\usepackage{algorithm}
\usepackage{algorithmic}
\usepackage{amsmath}
\usepackage{multirow}
\usepackage{graphicx}
\usepackage{booktabs} 
\usepackage{amsfonts}
\usepackage{amssymb}

\usepackage{newfloat}
\usepackage{listings}
\DeclareCaptionStyle{ruled}{labelfont=normalfont,labelsep=colon,strut=off} 
\floatstyle{ruled}
\newfloat{listing}{tb}{lst}{}
\floatname{listing}{Listing}
\title{Conditional Variational Autoencoder for Sign Language Translation with Cross-Modal Alignment}
\author{
    Rui Zhao, \textsuperscript{\rm 1,2}{\equalcontrib}
    Liang Zhang, \textsuperscript{\rm 1,2}{\equalcontrib}
    Biao Fu, \textsuperscript{\rm 1,2} 
    Cong Hu, \textsuperscript{\rm 1,2}
    Jinsong Su, \textsuperscript{\rm 1,2}
    Yidong Chen \textsuperscript{\rm 1,2}\thanks{Corresponding Author.}
}
\affiliations{
    \textsuperscript{\rm 1}School of Informatics, Xiamen University, China\\
    \textsuperscript{\rm 2}Key Laboratory of Digital Protection and Intelligent Processing of Intangible Cultural Heritage of \\ Fujian and Taiwan (Xiamen University), Ministry of Culture and Tourism, China \\    

    \{zhsqzr, lzhang\}@stu.xmu.edu.cn, ydchen@xmu.edu.cn
}

\usepackage{bibentry}

\begin{document}

\maketitle

\begin{abstract}
Sign language translation (SLT) aims to convert continuous sign language videos into textual sentences. As a typical multi-modal task, there exists an inherent modality gap between sign language videos and spoken language text, which makes the cross-modal alignment between visual and textual modalities crucial. However, previous studies tend to rely on an intermediate sign gloss representation to help alleviate the cross-modal problem thereby neglecting the alignment across modalities that may lead to compromised results.
To address this issue, we propose a novel framework based on \textbf{C}onditional \textbf{V}ariational autoencoder for \textbf{SLT} (\textbf{CV-SLT}) that facilitates direct and sufficient cross-modal alignment between sign language videos and spoken language text. Specifically, our CV-SLT consists of two paths with two Kullback-Leibler (KL) divergences to regularize the outputs of the encoder and decoder, respectively. In the \textit{prior path}, the model solely relies on visual information to predict the target text; whereas in the \textit{posterior path}, it simultaneously encodes visual information and textual knowledge to reconstruct the target text. The first KL divergence optimizes the conditional variational autoencoder and regularizes the encoder outputs, while the second KL divergence performs a self-distillation from the posterior path to the prior path, ensuring the consistency of decoder outputs.
We further enhance the integration of textual information to the posterior path by employing a shared Attention Residual Gaussian Distribution (ARGD), which considers the textual information in the posterior path as a residual component relative to the prior path.
Extensive experiments conducted on public datasets (PHOENIX14T and CSL-daily) demonstrate the effectiveness of our framework, achieving new state-of-the-art results while significantly alleviating the cross-modal representation discrepancy. The code and models are available at https://github.com/rzhao-zhsq/CV-SLT.
\end{abstract}

\section{Intruduction}

\begin{figure}[ht]
\centering
\includegraphics[width=0.45 \textwidth]{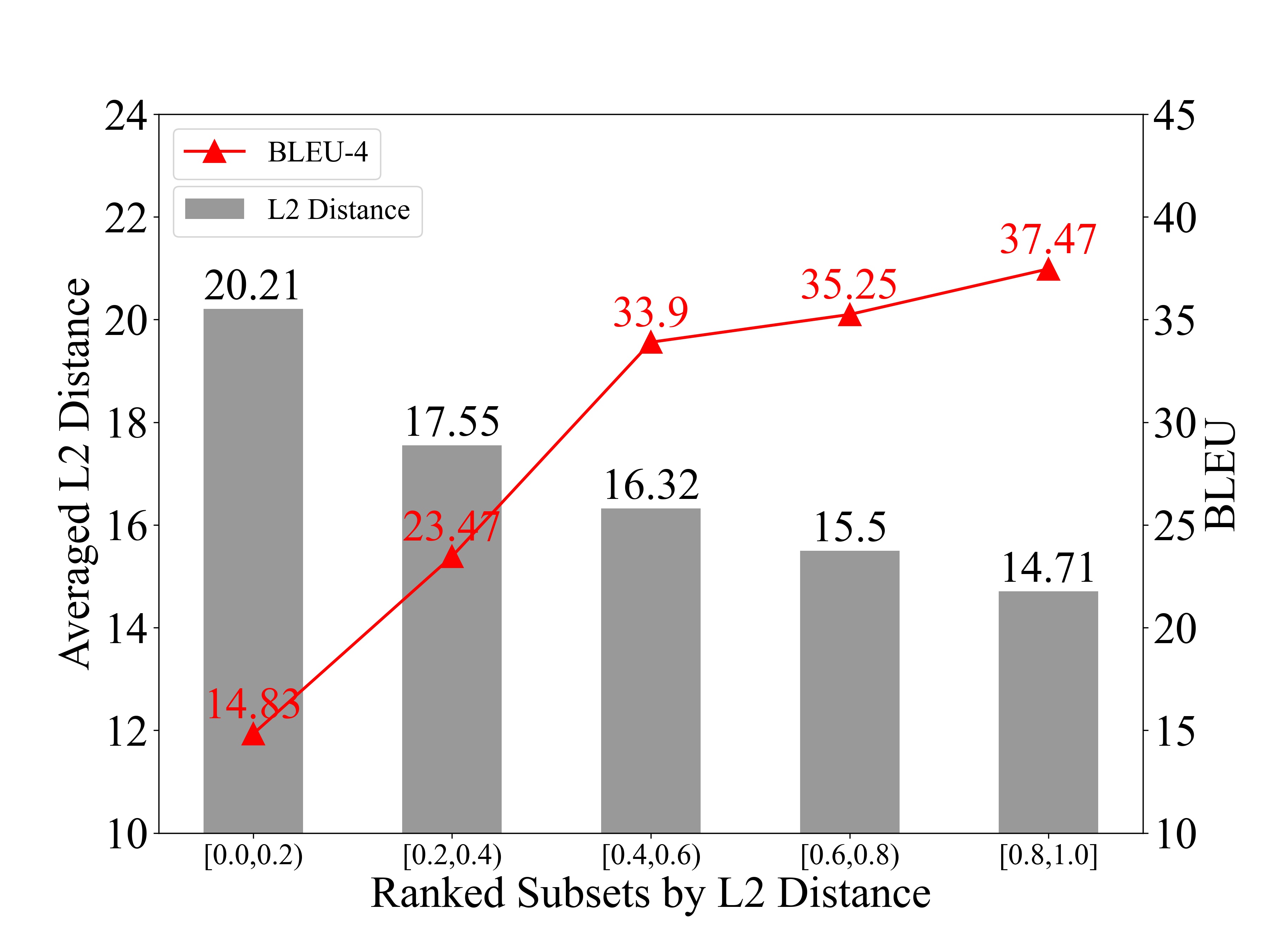} 
\caption{The PHOENIX14T Dev set is divided into 5 subsets of equal size according to the L2 distance of sentence-level representation between sign language and text embedding generated by previous state-of-the-art method MMTLB \cite{chenSimpleMultiModalityTransfer2022}. The histogram represents the L2 distance of each subset. Red triangle represents the BLEU-4 scores.
With the reduction of modality gap, a distinct upward trend in BLEU score is observed.
}
\label{fig_intro_1}
\end{figure}

Sign language serves as the primary mode of communication within the deaf community. It possesses a unique grammatical structure and lexicon that conveys semantic meanings through coordinated movements of the torso, head, hands, and other body parts. 
This characteristic distinguishes sign language from spoken language and poses significant barriers for hearing individuals to understand and communicate with deaf people.
Sign language translation (SLT), which aims to translate sign language videos into spoken language text, has gained increasing attention as a means to bridge the communication gap between hearing-impaired and unimpaired individuals 
\cite{camgozNeuralSignLanguage2018,
cihancamgozSignLanguageTransformers2020,
kamalTechnicalApproachesChinese2019,
yin-etal-2021-including,
chenSimpleMultiModalityTransfer2022,
chenTwoStreamNetworkSign2022,
fuTokenLevelContrastiveFramework2023,
yuEfficientSignLanguage2023}.

Existing SLT methods adopt the neural machine translation (NMT) paradigm, treating SLT as a sequence-to-sequence task. 
However, unlike NMT where both the source and the targets are in textual modality, SLT involves sign language videos as the source and requires translation into textual sentences, thereby creating a tough challenge due to the inherent modality gap.
Therefore, some conventional approaches \cite{camgozNeuralSignLanguage2018,yinBetterSignLanguage2020,zhouImprovingSignLanguage2021} adopt a sequential pipeline framework known as Sign2Gloss2Text, which first employs a Sign2Gloss module to map sign language videos to the corresponding gloss\footnote{Glosses are word-by-word spoken language textual words that approximately match the meaning of sign language.} sequences and then utilizes a Gloss2Text module to translate the recognized glosses into the final text. 
However, training the Sign2Gloss and Gloss2Text modules independently may result in a lack of alignment between sign language videos and target text while the cascaded pipeline may suffer from error propagation. 
To overcome these limitations, some recent methods \cite{cihancamgozSignLanguageTransformers2020, chenTwoStreamNetworkSign2022,zhangSLTUNETSimpleUnified2023} propose building a unified model for jointly learning sign language recognition (SLR) and SLT.
Though these multi-task joint learning methods have achieved improved performance, the alignment between visual and textual modalities remains insufficient.
As illustrated in Fig. \ref{fig_intro_1}, we divide the PHOENIX14T Dev set into five subsets of equal size according to the L2 distance between paired sentence-level sign language representation and text embedding which are generated by previous state-of-the-art method MMTLB \cite{chenSimpleMultiModalityTransfer2022}. 
It is observed that the subsets in sections [0.0,0.2) and [0.2,0.4) exhibit higher L2 distances, indicating the presence of a noticeable modality gap that leads to significantly degraded BLUE scores. However, as the L2 distance decreases, there is a distinct upward trend in BLEU scores.

Therefore, to directly and sufficiently align the visual and textual modalities, we propose a novel framework based on \textbf{C}onditional \textbf{V}ariational autoencoder for \textbf{SLT} (CV-SLT) that includes two paths: \textit{prior path} and \textit{posterior path}. 
In the prior path, the model only relies on the information of visual modality to predict the target text, while in the posterior path, the model simultaneously encodes visual information and textual knowledge to reconstruct the target text. Notably, gloss is not required in either path. 
We use two Kullback-Leibler (KL) divergences to facilitate alignment between visual and textual modalities.
The first KL divergence optimizes the conditional variational autoencoder (CVAE) and aligns
the encoder outputs across both paths,
closing the obtained uni-modal marginal distribution of prior encoder outputs with the bimodal joint distribution of posterior encoder outputs, to bridge the modality gap. 
The second KL divergence performs a self-distillation from the posterior path to the prior path, thereby ensuring the alignment and consistency of the decoder's outputs, regardless of whether the input comprises uni-modal information from sign language videos or bimodal information from both sign language videos and text.

Furthermore, to enhance the integration of textual information into the posterior path and address the length discrepancy between sign language videos and target text, we employ a shared Attention Residual Gaussian Distribution (ARGD) for the posterior path. The ARGD models the posterior distribution using relative location and scale instead of absolute ones through a shared attention mechanism. Specifically, self-attention is utilized for uni-modal visual information in the prior path, while cross-attention is applied in the posterior path with visual information as query and textual information as key/value pairs, resulting in a residual Gaussian distribution relative to the prior.

Our main contributions are as follows:
\begin{itemize}

\item We propose CV-SLT, which consists of two paths with two Kullback-Leibler (KL) divergences, aiming to alleviate the modality gap in sign language translation. To our knowledge, this is the first application of CVAE in SLT.

\item A shared Attention Residual Gaussian Distribution (ARGD) is utilized to further enhance integrating textual information to the posterior distribution, which considers the textual information as a residual component relative to the prior distribution.

\item Extensive experiments conducted on PHOENIX14T and CSL-daily datasets demonstrate that our model significantly outperformed strong baselines, achieving new state-of-the-art in SLT while mitigating the desire for gloss, which we believe is a promising evolution.

\end{itemize}

\begin{figure*}[ht]
\centering
\includegraphics[width=0.9 \textwidth]{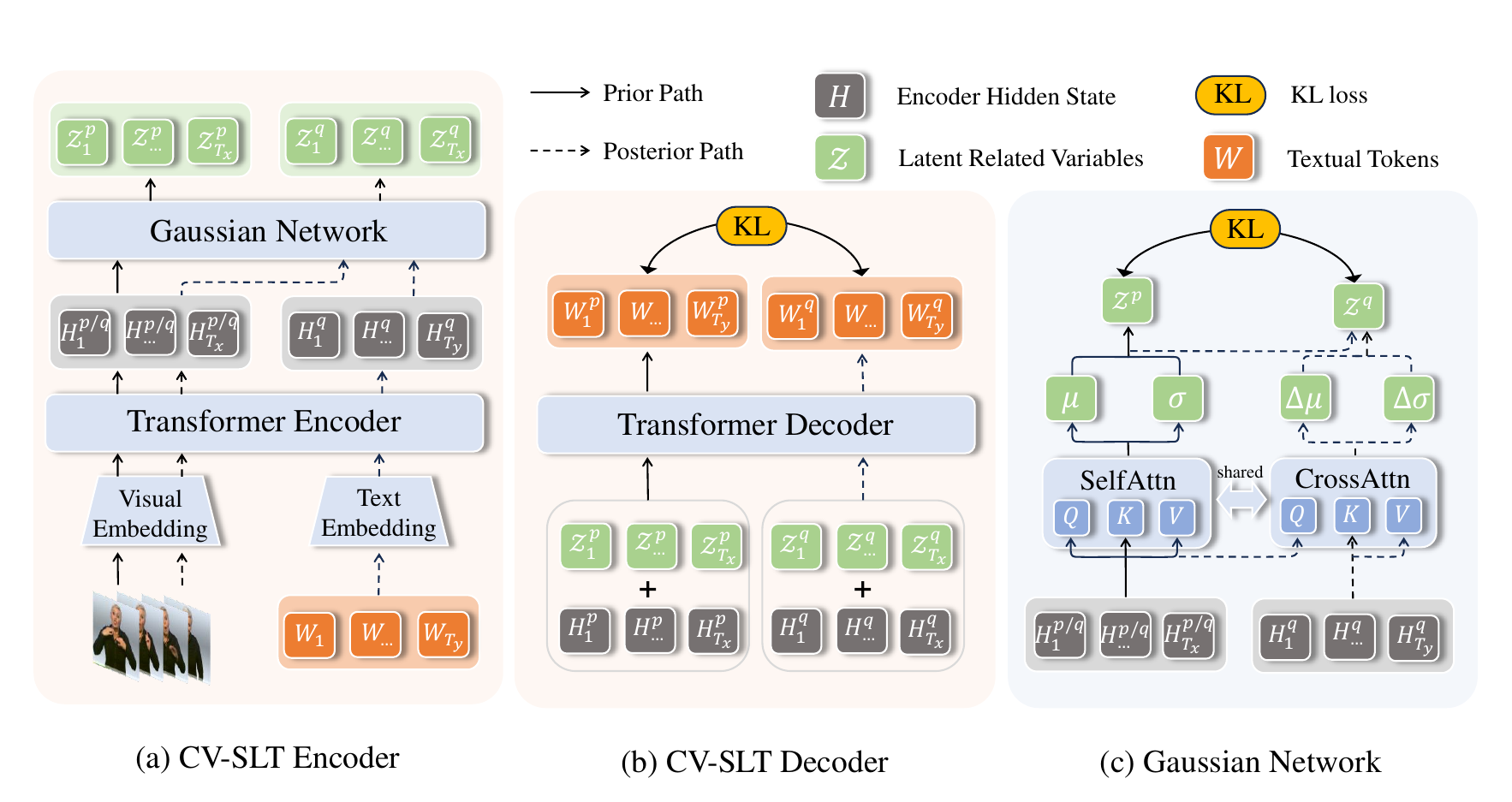} 
\caption{Detailed model framework of our CV-SLT, which adopts an encoder-decoder architecture. The Gaussian Network models the posterior $q_{\phi}(z|x,y)$ relative to the  $p_{\theta}(z|x)$ with shared Attention Residual Gaussian Distribution (ARGD). The encoder and decoder outputs from the prior (solid line) and posterior (dashed line) paths are respectively regularized with two KL divergences, as in (b) and (c). During inference, only the prior path is employed.}
\label{detailed_model}
\end{figure*}

\section{Methods}

\subsection{CVAE-based SLT} \label{methods_1}
Supposing that $x=\{x_1,...,x_{T_x}\}$ and $y=\{y_1,...,y_{T_y}\}$ represent sign language videos and corresponding spoken language text with length $T_x$ and $T_y$, respectively, we introduce latent variables $z=\{z_1,...,z_{T_x}\}$ to help model the conditional probability of SLT: 
\begin{equation}
p\left( y|x \right) =\int_z{p\left( y,z|x \right) dz}=\int_z{p\left( y|z,x \right) p\left( z|x \right) dz}.
\end{equation}
In this way, the latent variables can serve as a springboard to help bridge the visual and textual modalities. Specifically, the latent variables drawn from the prior distribution $p_{\theta}(z|x)$ could be viewed as a uni-modal semantic representation of sign language, while the posterior latent variables drawn from the posterior distribution $p_{\theta}(z|x,y)$ contain a bimodal semantic representation of both sign language and text. The model is trained using a variational approximate posterior distribution, denoted as $q_{\phi}(z|x,y)$, due to the intractability of the true posterior distribution $p_{\theta}(z|x,y)$. The objective is to optimize the Evidence Lower Bound (ELBO):
\begin{equation} \label{cvae_elbo}
\begin{aligned}
\mathcal{L}_{\text{CVAE}}\left( \theta, \phi; x, y\right) = 
- \operatorname{KL}\left( q_{\phi}\left( z|x,y \right) ||p_{\theta}\left( z | x \right) \right)  \\
+ E_{q_{\phi}\left( z|x,y \right)}\left[ \log p_{\theta}\left( y|z,x \right) \right] 
\le \log p_{\theta}\left( y|x \right).
\end{aligned}
\end{equation}
By leveraging approximate posterior inference and reparameterization technique, the prior can effectively capture bimodal information from the posterior distribution, thereby facilitating cross-modal alignment.

\begin{table*}[ht]
\centering
\small
\renewcommand{\arraystretch}{1.0}
\resizebox{0.9 \linewidth}{!}
{
\begin{tabular}{lcccccccccc}

\toprule

\multicolumn{1}{c|}{\multirow{2}{*}{\textbf{Methods}}} & \multicolumn{5}{c|}{\textbf{Dev}} & \multicolumn{5}{c}{\textbf{Test}} \\
\multicolumn{1}{c|}{} & \textbf{R} & \textbf{B1} & \textbf{B2} & \textbf{B3} & \multicolumn{1}{c|}{\textbf{B4}} & \textbf{R} & \textbf{B1} & \textbf{B2} & \textbf{B3} & \textbf{B4} \\ 
\cmidrule[0.6pt](lr){1-11}
\multicolumn{11}{c}{\textbf{PHOENIX14T}} \\ 

\cmidrule(lr){1-11}

\multicolumn{1}{l|}{SL-Luong \cite{camgozNeuralSignLanguage2018}} & 31.80 & 31.87 & 19.11 & 13.16 & \multicolumn{1}{c|}{9.94} & 31.80 & 32.24 & 19.03 & 12.83 & 9.58 \\
\multicolumn{1}{l|}{TSPNet \cite{liTSPNetHierarchicalFeature2020}} & - & - & - & - & \multicolumn{1}{c|}{-} & 34.96 & 36.10 & 23.12 & 16.88 & 13.41 \\
\multicolumn{1}{l|}{Joint-SLRT \cite{camgozMultichannelTransformersMultiarticulatory2020}} & - & 45.54 & 32.60 & 25.30 & \multicolumn{1}{c|}{20.69} & - & 45.34 & 32.31 & 24.83 & 20.17 \\
\multicolumn{1}{l|}{MMTLB \cite{chenSimpleMultiModalityTransfer2022}} &  53.06 & 53.33 &  40.67 &  32.75 & \multicolumn{1}{c|}{ 27.25} &  52.44 &  53.41 &  41.08 & 33.27 & 27.95 \\ 
\cmidrule(lr){1-11}
\multicolumn{1}{l|}{Ours} & $\textbf{54.43}^{\dag}$ & $\textbf{55.09}^{\ddag}$ & $\textbf{42.60}^{\ddag}$ & $\textbf{34.63}^{\ddag}$ & \multicolumn{1}{c|}{$\textbf{29.10}^{\ddag}$} & $\textbf{54.33}^{\ddag}$ & $\textbf{54.88}^{\dag}$ & $\textbf{42.68}^{\ddag}$ & $\textbf{34.79}^{\dag}$ & $\textbf{29.27}^{\dag}$ \\
\multicolumn{1}{l|}{Improvement} & +1.37 & +1.76 & +1.93 & +1.88 & \multicolumn{1}{c|}{+1.85} & +1.89 & +1.47 & +1.60 & +1.52 & +1.32 \\ 

\midrule

\multicolumn{11}{c}{\textbf{CSL-daily}} \\ 
\cmidrule[0.6pt](lr){1-11}
\multicolumn{1}{l|}{SL-Luong \cite{camgozNeuralSignLanguage2018}} & 34.28 & 34.22 & 19.72 & 12.24 & \multicolumn{1}{c|}{7.96} & 34.54 & 34.16 & 19.57 & 11.84 & 7.56 \\
\multicolumn{1}{l|}{Joint-SLRT \cite{camgozMultichannelTransformersMultiarticulatory2020}} & 26.75 & 28.81 & 16.48 & 10.78 & \multicolumn{1}{c|}{7.59} & 26.91 & 28.70 & 16.33 & 10.52 & 7.29 \\
\multicolumn{1}{l|}{ MMTLB \cite{chenSimpleMultiModalityTransfer2022}} & 55.51 & 57.13 & 43.81 & 34.35 & \multicolumn{1}{c|}{27.57} & 55.40 & 56.97 & 43.70 & 34.14 & 27.30 \\ 
\cmidrule(lr){1-11}
\multicolumn{1}{l|}{\textbf{Ours}} & \textbf{56.36} & \textbf{58.05} & \textbf{44.73} & \textbf{35.14} & \multicolumn{1}{c|}{\textbf{28.24}} & \textbf{57.06} & $\textbf{58.29}^{\dag}$ & $\textbf{45.15}^{\ddag}$ & $\textbf{35.77}^{\ddag}$ & $\textbf{28.94}^{\ddag}$ \\
\multicolumn{1}{l|}{Improvement} & +0.85 & +0.92 & +0.92 & +0.79 & \multicolumn{1}{c|}{+0.67} & +1.66 & +1.32 & +1.45 & +1.63 & +1.64 \\ 
\bottomrule
\end{tabular}

}

\caption{Brief experimental results on PHOENIX14T (upper) and CSL-daily (bottom) compared with baselines which excludes gloss supervision during SLT training. B-n means BLEU-n, R represents ROUGE. Our CV-SLT simultaneously outperforms all baselines by a significant margin on both datasets. The optimal results are highlighted in \textbf{bold}. The performance improvement is also displayed for clear comparison.
``$\dag$" and ``$\ddag$" indicate the improvement over MMTLB is statistically significant ($p<0.05$ and $p<0.01$, respectively), estimated by bootstrap sampling \cite{koehnStatisticalSignificanceTests2004}.
}
\label{tab_main_results}
\end{table*}

\subsection{Model Details} \label{methods_2}

As shown in Fig. \ref{detailed_model}(a), to 
align the representation of encoder outputs, we first need to incorporate the textual information into the approximate posterior $q_{\phi}(z|x,y)$. Specifically, we input the visual feature into the Transformer-Encoder twice with and without text respectively to get the corresponding visual and text representation for posterior $q_\phi$ and prior $p_{\theta}$:
\begin{equation} \label{posterior_encoder_out}
H^q_{\text{Vison}}, H^q_{\text{Text}} = \text{Encoder}([x;y]), 
\end{equation} 
\begin{equation}
H^p_{\text{Vison}} = \text{Encoder}(x),
\end{equation}
then, the prior and approximate posterior distributions are modeled using a Gaussian network with ARGD, enabling the derivation of latent variables for both paths.

\subsubsection{ARGD for Posterior Path.}
As shown in Fig. \ref{detailed_model}(c), we first employ an attention mechanism that shares weights across both paths to map the obtained encoder outputs to a shared semantic space since the source sign language and target text are usually semantically equivalent:
\begin{equation} \label{posterior_attn}
H^p = \text{SelfAttn}(H^p_{\text{Vison}}, H^p_{\text{Vison}}, H^p_{\text{Vison}}),
\end{equation}
\begin{equation}
H^q = \text{CrossAttn}(H^q_{\text{Vison}}, H^q_{\text{Text}}, H^q_{\text{Text}}),
\end{equation}
where $\text{Attn}(Q, K, V)$ is dot-product attention. 
$\text{SelfAttn}(\cdot)$ and $\text{CrossAttn}(\cdot)$ are the same as that in Transformer \cite{vaswaniAttentionAllYou2017a}. 

Then, we parameterize the prior as a multivariate Gaussian distribution  and employ a linear network $f(\cdot)$ to calculate the pivotal $d_{z}$-dimension vectors $\mu$ and $\sigma$ for $z$, parameterized as $W^f_{\mu}, W^f_{\sigma} \in \mathbb{R}^{d_{k} \times d_{z}}$:
\begin{equation}
    p_{\theta}(z|x) = N(\mu, \text{diag}(\sigma^2)),
\end{equation}
\begin{equation}
[\mu ,\ \log\sigma^2] = f(H^p).
\end{equation}
With the help of reparameterization technique, we get the prior latent variables with:
\begin{equation}
z=\mu + \sigma \odot \epsilon,
\end{equation}
where $\epsilon$ is a standard Gaussian noise and $\odot$ denote an element-wise product.

For posterior distribution, we still model it as a multivariate Gaussian distribution but utilize a residual distribution to parameterize the $q_{\phi}(z|x,y)$ related to $p_{\theta}(z|x)$:
\begin{equation}
q_{\phi}(z|x,y) = N(\mu + \Delta\mu, \text{diag}(\sigma^2 \cdot \Delta\sigma^2) ),
\end{equation}
\begin{equation} \label{residual_predict}
[\Delta\mu,\log\Delta\sigma^2 ] = g(H^q), 
\end{equation}
where $\Delta\mu$ and $\Delta\sigma$ are relative location and scale of the approximate posterior with respect to the prior. Here the linear layer $g(\cdot)$ parameterized as $W^g_{\Delta\mu}, W^g_{\Delta\sigma} \in \mathbb{R}^{d \times n}$ is different from that in prior. With the same noise sampled from standard Gaussian noise and reparameterization, the KL term of ELBO in Eq.(\ref{cvae_elbo}) could be written as:
\begin{equation} \label{residual_kl}
\text{KL} =\frac{1}{2}\left( \frac{\Delta \mu ^2}{\sigma ^2}+\Delta \sigma ^2-\log \Delta \sigma ^2-1 \right). 
\end{equation}

Finally, it is trained to optimize the objective $ \mathcal{L}_{\text{CVAE}}\left( \theta, \phi; x, y\right) $ described in 
Eq.(\ref{cvae_elbo}) for the posterior path.

Modelling the posterior path via our ARGD offers several salient benefits.
Firstly, in the attention mechanism, we can explicitly incorporate extra text information into the corresponding visual part via cross attention, which works as a compensation for implicitly incorporation in Eq.(\ref{posterior_encoder_out})
Secondly, intuitively, a parameter-shared attention for both posterior and prior can map sign language and text to a unified representation space compared with a parameter-independent attention, facilitating to further shrink the gap between visual and textual modalities. 
Empirical analyses in Sec.\ref{sec_component_study} also demonstrate the effectiveness of the parameter-shared attention.
Finally, minimizing the KL term in the residual parameterization is easier than when $q_{\phi}(z|x,y)$ predict the absolute location and scale \cite{vahdatNVAEDeepHierarchical2020,huRecurrenceBoostsDiversity2022}, and, the formulation also contributes to 
stabilizing the training and mitigating the KL vanishing problem \cite{bowmanGeneratingSentencesContinuous2016}.

\subsubsection{Alignment Enhanced Prior Path.} \label{Sec_AEP}
For CVAE the $q_{\phi}(z|x,y)$ is used at training but at inference, the $p_{\theta}(z|x)$ is employed to draw variables $z$ and to make a prediction. It is much easier for the decoder to predict $y$ since during training, $y$ is given as the input for the encoder and the objective can be viewed as a reconstruction of $y$. Though the KL term in Eq.(\ref{cvae_elbo}) is to close the gap between two pipelines, the discrepancy is still intractable due to the significant modality-gap.

Following \citeauthor{sohnLearningStructuredOutput2015} \shortcite{sohnLearningStructuredOutput2015}, we train the networks in a way that the prediction pipelines at training and inference are consistent. To achieve this end, we utilize the latent variables sampled from prior net $p_{\theta}(z|x)$ and predict the $y$ together with $x$:
\begin{equation} \label{AEP_loss}
\mathcal{L}_{\text{AEP}}\left( \theta; x, y\right) = E_{p_{\theta}\left( z|x \right)}\left[ \log p_{\theta}\left( y|z,x \right) \right],
\end{equation}
where $z$ are the prior latent variables. 

Subsequently, we employ an additional Kullback-Leibler divergence to facilitate Self-Distillation (SD) between the prior and posterior paths, thereby promoting the prior to learn cross-modal knowledge from the posterior:
\begin{equation} \label{loss_SD}
    \mathcal{L}_{\text{SD}} = \operatorname{KL}(p_{\theta}(y^{q}|x,z) ||p_{\theta}(y^{p}|x,z) ),
\end{equation}
where $y^{q}$ and $y^{p}$ are predictions from the posterior and prior paths respectively. This process serves to regularize the encoder outputs while simultaneously enhancing the alignment between visual and textual modalities.

\subsection{Training and Inference} \label{methods_3}
\subsubsection{Training.}
During, we combine the objective functions of two paths to obtain a hybrid objective:
\begin{equation} \label{Total_loss}
    \mathcal{L} = \mathcal{L}_{\text{CVAE}} + \mathcal{L}_{\text{AEP}} + \lambda \mathcal{L}_{\text{SD}}.
\end{equation} 
Here we set $\lambda$ as a hyper-parameter which controls the regularization from posterior to prior.

\subsubsection{Inference.} Once the model parameters are well trained, we can make a prediction conditioned on $x$ with the prior path. However, the standard Gaussian noise $\epsilon$ introduced in prior latent variables might lead to ambiguity. As an alternative approach, we perform a deterministic inference without sampling $z$:
\begin{equation}
y = \underset{y}{\arg \, \max} \ p_{\theta}(y|x,z^{*}), z^{*} = E\left[ z|x \right].
\end{equation}

Although there exist more advanced approaches to evaluate the conditional likelihood during inference, such as Monte Carlo sampling and importance sampling \cite{burdaImportanceWeightedAutoencoders2016},  their utilization in this work is constrained due to computational limitations.

\begin{table*}[ht]
\centering
\small
\renewcommand{\arraystretch}{1.0}
\resizebox{0.9 \linewidth}{!}{
\begin{tabular}{l|cc|ccccc|c}
\toprule
\multirow{2}{*}{Methods} & \multicolumn{2}{c|}{Dev} & \multicolumn{5}{c|}{Test} & \multirow{2}{*}{Group} \\
 & R & B4 & R & B1 & B2 & B3 & B4 &  \\ 
 \cmidrule(lr){1-9}
TSPNet \cite{liTSPNetHierarchicalFeature2020} & - & - & 34.96 & 36.10 & 23.12 & 16.88 & 13.41 &  \\
SL-Luong \cite{camgozNeuralSignLanguage2018} & 44.14 & 18.40 & 43.80 & 43.29 & 30.39 & 22.82 & 18.13 & \multirow{4}{*}{Pipeline} \\
Joint-SLRT \cite{camgozMultichannelTransformersMultiarticulatory2020} & - & 22.11 & - & 48.47 & 35.35 & 27.57 & 22.45 &  \\
SignBT \cite{zhouImprovingSignLanguage2021} & 49.53 & 23.51 & 49.35 & 48.55 & 36.13 & 28.47 & 23.51 &  \\
SMTC-T \cite{yinBetterSignLanguage2020} & 48.70 & 24.68 & 48.78 & 50.63 & 38.36 & 30.58 & 25.40 &  \\
\cmidrule(lr){1-9}
Joint-SLRT \cite{camgozMultichannelTransformersMultiarticulatory2020} & - & 22.38 & - & 46.61 & 33.73 & 26.19 & 21.32 & \multirow{4}{*}{Jointly} \\
MMTLB \cite{chenSimpleMultiModalityTransfer2022} & 53.10 & 27.61 & 52.65 & 53.97 & 41.75 & 33.84 & 28.39 &  \\
SLTUNET \cite{zhangSLTUNETSimpleUnified2023} & 52.23 & 27.87 & 52.11 & 52.92 & 41.76 & 33.99 & 28.47 &  \\
TS-SLT \cite{chenTwoStreamNetworkSign2022} & 54.08 & 28.66 & 53.48 & \textbf{54.90} & 42.43 & 34.46 & 28.95 &  \\ 
\cmidrule(lr){1-9}
CV-SLT(Ours) & \textbf{54.43} & \textbf{29.10} & \textbf{54.33} & 54.88 & \textbf{42.68} & \textbf{34.79} & \textbf{29.27} & Variational \\ 
\bottomrule
\end{tabular}
}
\caption{
Compare with state-of-the-art methods on PHOENIX14T. Our CV-SLT aligns the visual and textual modalities directly with CVAE, achieving a significant improvement despite the absence of gloss supervision during the SLT training, which shows the great potential of variational alignment for SLT.
}
\label{tab_sota_phoenix}
\end{table*}

\section{Experiments}

\subsection{Datasets and Evaluation Metrics}

\subsubsection{Datasets.}

To evaluate the effectiveness of our proposed CV-SLT, we conduct extensive experiments on the following publicly available datasets:
\begin{itemize}
    \item PHOENIX14T \cite{camgozNeuralSignLanguage2018}: PHOENIX14T is the most widely used benchmark for SLT in recent years. It contains 8,257 parallel German sign language videos with German translations from weather forecast programs, split into Train/Dev and Test sets of sizes 7,096/519 and 642, respectively.
    \item CSL-daily \cite{zhouImprovingSignLanguage2021}: CSL-Daily focuses on daily topics in Chinese sign language, containing 20,654 parallel CSL videos with Chinese translations. The dataset is split into Train/Dev and Test sets of sizes 18,401/1,077 and 1,176, respectively.
\end{itemize}

\subsubsection{Evaluation Metrics.} 
Following previous works \cite{chenSimpleMultiModalityTransfer2022,chenTwoStreamNetworkSign2022}
, we adopt BLEU \cite{papineniBleuMethodAutomatic2002a} and ROUGE \cite{linROUGEPackageAutomatic2004a} to evaluate frameworks for SLT. Higher BLEU and ROUGE indicate better translation performance.

\subsection{Implementation and Optimization Details} 
Following MMTLB \cite{chenSimpleMultiModalityTransfer2022}, We use the same configuration for visual embedding and pretrained mBart encoder-decoder. Besides, The Gaussian Network consists of a one-layer pure attention mechanism without any additional feedforward network (FFN) layers nor residual connection, while the prior and posterior are both one-layer linear layers.
We adopt a learning rate of $1e-5$ and select 64 for the dimension ($d_{z}$) of latent variables. The self-distillation weight ($\lambda$) is set to 3 according to preliminary experiments. 
The KL annealing trick \cite{bowmanGeneratingSentencesContinuous2016} is used to avoid KL vanishing during training for the first 4K steps.
During inference, we follow previous studies \cite{chenSimpleMultiModalityTransfer2022,chenTwoStreamNetworkSign2022} to use beam search with a length penalty of 1 and a beam size of 5.
The batch size is set to 16 and AMP \cite{baboulinAcceleratingScientificComputations2009} is applied due to the computation limitation. 
We implement our CV-SLT based on open-source SLRT\footnote{\url{https://github.com/FangyunWei/SLRT}}. All experiments are conducted on a single NVIDIA TITAN RTX GPU.

\subsection{Main Results}
\begin{table*}[ht]
\centering
\small
\renewcommand{\arraystretch}{1.0}
\resizebox{0.9 \linewidth}{!}{
\begin{tabular}{l|cc|ccccc|c}
\toprule
\multirow{2}{*}{Methods} & \multicolumn{2}{c|}{Dev} & \multicolumn{5}{c|}{Test} & \multirow{2}{*}{Group} \\
 & R & B4 & R & B1 & B2 & B3 & B4 &  \\ 
\cmidrule(lr){1-9}
SL-Luong \cite{camgozNeuralSignLanguage2018} & 40.18 & 11.06 & 40.05 & 41.55 & 25.73 & 16.54 & 11.03 & \multirow{3}{*}{Pipeline} \\
Joint-SLRT \cite{camgozMultichannelTransformersMultiarticulatory2020} & 37.06 & 11.88 & 36.74 & 37.38 & 24.36 & 16.55 & 11.79 &  \\
SignBT \cite{zhouImprovingSignLanguage2021} & 48.38 & 19.53 & 48.21 & 50.68 & 36.00 & 26.20 & 19.67 &  \\ 
\cmidrule(lr){1-9}
Joint-SLRT \cite{camgozMultichannelTransformersMultiarticulatory2020} & 44.18 & 15.94 & 44.81 & 47.09 & 32.49 & 22.61 & 16.24 & \multirow{5}{*}{Jointly} \\
MMTLB \cite{chenSimpleMultiModalityTransfer2022} & 53.38 & 24.42 & 53.25 & 53.31 & 40.41 & 30.87 & 23.92 &  \\
SLTUNET \cite{zhangSLTUNETSimpleUnified2023} & 53.58 & 23.99 & 54.08 & 54.98 & 41.44 & 31.84 & 25.01 &  \\
TS-SLT \cite{chenTwoStreamNetworkSign2022} & 55.10 & 25.76 & 55.72 & 55.44 & 42.59 & 32.87 & 25.79 &  \\
MMTLB-fixed & 56.10 & 27.53 & 55.81 & 56.27 & 43.24 & 34.07 & 27.46 & \\
\cmidrule(lr){1-9}
CV-SLT (Ours) & \textbf{56.36} & \textbf{28.24} & \textbf{57.06} & \textbf{58.29} & \textbf{45.15} & \textbf{35.77} & \textbf{28.94} & Variational \\ 
\bottomrule
\end{tabular}
}
\caption{
Comparison with state-of-the-art methods on CSL-daily. Our CV-SLT outperforms all previous methods consistently. A bug is identified in MMTLB where it fails to handle the conversion between full Angle and half Angle in Chinese corpus. We have addressed this issue and re-evaluated MMTLB on CSL-daily using their proposed checkpoint.
}
\label{tab_sota_csl}
\end{table*}

\subsubsection{Comparison with Baselines without Gloss Supervision.}
In this work, no additional gloss supervision was introduced during the SLT training. 
For a fair comparison, we first evaluate our model against several methods with an identical setup on the PHOENIX14T and CSL-daily datasets in Table \ref{tab_main_results}.
For MMTLB, we reproduce and report its results by setting the SLR-related loss weight to $0$ for a fair comparison since they jointly learn SLR and SLT, as well as the result for Joint-SLRT on CSL-daily since the original paper only focuses on PHOENIX14T. The performance of SL-Luong on CSL-daily is from \cite{zhouImprovingSignLanguage2021}. Other results are reported from the original paper.
As illustrated in Table \ref{tab_main_results}, the translation performance of our model simultaneously outperforms all baselines by a significant margin (+1.85/+1.32 BLEU and +0.67/+1.64 BLEU) on Dev/Test sets of PHOENIX14T and CSL-daily, respectively. 
Due to the cross-modal alignment ability of our CV-SLT, the visual information of sign language and textual information of target text are mapped to a similar semantic space through an aligned encoder, thereby empowering the decoder to accurately predict the target text.

\subsubsection{Comparison with State-of-the-art Methods.}
We further compare our CV-SLT to the current state-of-the-art methods on PHOENIX14T and CSL-daily datasets. To begin with, We divide previous SOTA methods into two groups according to how they deal with the representation discrepancy of sign language videos and spoken language text, 1) \textbf{Pipeline}: these methods perform Sign2Gloss and Gloss2Text in a cascaded manner, 
2) \textbf{Jointly}: these methods learn SLR and SLT jointly.

The performance of our CV-SLT consistently outperforms all previous state-of-the-art methods, as illustrated in Table \ref{tab_sota_phoenix} and Table \ref{tab_sota_csl}, with an improvement of +0.44/+0.32 and +0.71/+1.48 BLEU scores on the Dev/Test sets of PHOENIX14T and CSL-daily.
Notably, despite the absence of gloss supervision during the SLT training, CV-SLT still achieves higher translation quality compared to SOTA methods that employ gloss supervision.
This improvement is attributed to effectively addressing the cross-modal representation issues by aligning visual and textual modalities. 
In the following sections, we will provide a detailed description.

\begin{table}[htb]
\centering
\small
\renewcommand{\arraystretch}{1.0}
\resizebox{0.9 \linewidth}{!}{
\begin{tabular}{l|cc|cc}
\toprule
\multirow{2}{*}{Methods} & \multicolumn{2}{c|}{Dev} & \multicolumn{2}{c}{Test} \\
 & R & B4 & R & B4 \\ 
\cmidrule(lr){1-5}
CV-SLT & \textbf{54.43} & \textbf{29.10} & \textbf{54.33} & \textbf{29.27} \\ 
\cmidrule(lr){1-5}
\quad w/o $\mathcal{L}_{\text{AEP}}$  & 27.40 & 4.74 & 26.12 & 5.34  \\
\quad w/o $\mathcal{L}_{\text{SD}}$ & 52.78 & 27.41 & 53.50 & 28.59 \\
\cmidrule(lr){1-5}
\quad w/o ARGD & 52.83 & 27.63 & 50.96 & 26.67 \\
\quad w/o Attention Shared &  53.98 &  28.59 & 53.59  &  29.24 \\ 
\bottomrule
\end{tabular}
}
\caption{\normalsize{Studies of contribution for each component on Dev/Test sets of PHOENIX14T.}}
\label{tab_ablation_main_component}
\end{table}

\subsection{Ablation Study}

\subsubsection{How each component contributes to our model?}
\label{sec_component_study}
To further demonstrate the contributions of different components of our CV-SLT, we conduct ablation studies on both Dev and Test sets of PHOENIX14T. To validate the Self-Distillation and AEP components of our CV-SLT described in Sec \ref{Sec_AEP}, we initially exclude the $\mathcal{L}_{\text{SD}}$ and $\mathcal{L}_{\text{AEP}}$ terms from our loss function in Eq.(\ref{Total_loss}). Subsequently, we replace the ARGD for posterior in the Gaussian network with a vanilla absolute posterior, where prior and posterior latent variables are modelled independently. Finally, to verify our insight into the attention mechanism, we set our attention for ARGD as non-shared.

As shown in Table \ref{tab_ablation_main_component}, the removal of $\mathcal{L}_{\text{AEP}}$ from our CV-SLT leads to a discrepancy between training and inference, resulting in a significant decline in both ROUGE and BLEU scores. We impute this collapse to the overfitting of the decoder known as KL vanishing, which has been discussed extensively in previous studies \cite{bowmanGeneratingSentencesContinuous2016, higginsBetaVAELearningBasic2017,zhuBatchNormalizedInference2020,shenRegularizingVariationalAutoencoder2021}. However, upon incorporating the entire prior path into our framework, our CV-SLT outperforms the baseline MMTLB \cite{chenSimpleMultiModalityTransfer2022} even without gloss supervision on the Test set (28.39 vs. 28.59). The inclusion of $\mathcal{L}_{\text{SD}}$ is a Self-Distillation mechanism that effectively regularizes the prior to learn from the posterior path and ensures consistency between prior and posterior distributions. With the incorporation of $\mathcal{L}_{\text{SD}}$, we can achieve further improvement with a BLEU score of 29.10/29.27 on Dev/Test sets.

In comparison to the vanilla variation without ARGD that the prior and posterior are modeled independently, our ARGD-version yields a notable enhancement of 1.47/2.60 BLUE scores on the Dev and Test sets, thus demonstrating the superior integration of textual information achieved by our ARGD. The shared attention is crucial as well since it not only yields an improvement of 0.51 BLUE scores on the Dev set but also effectively reduces the model parameters.

\subsubsection{What's the relationship between prior and posterior?}

\begin{figure}[t]
\centering
\includegraphics[width=0.4 \textwidth]{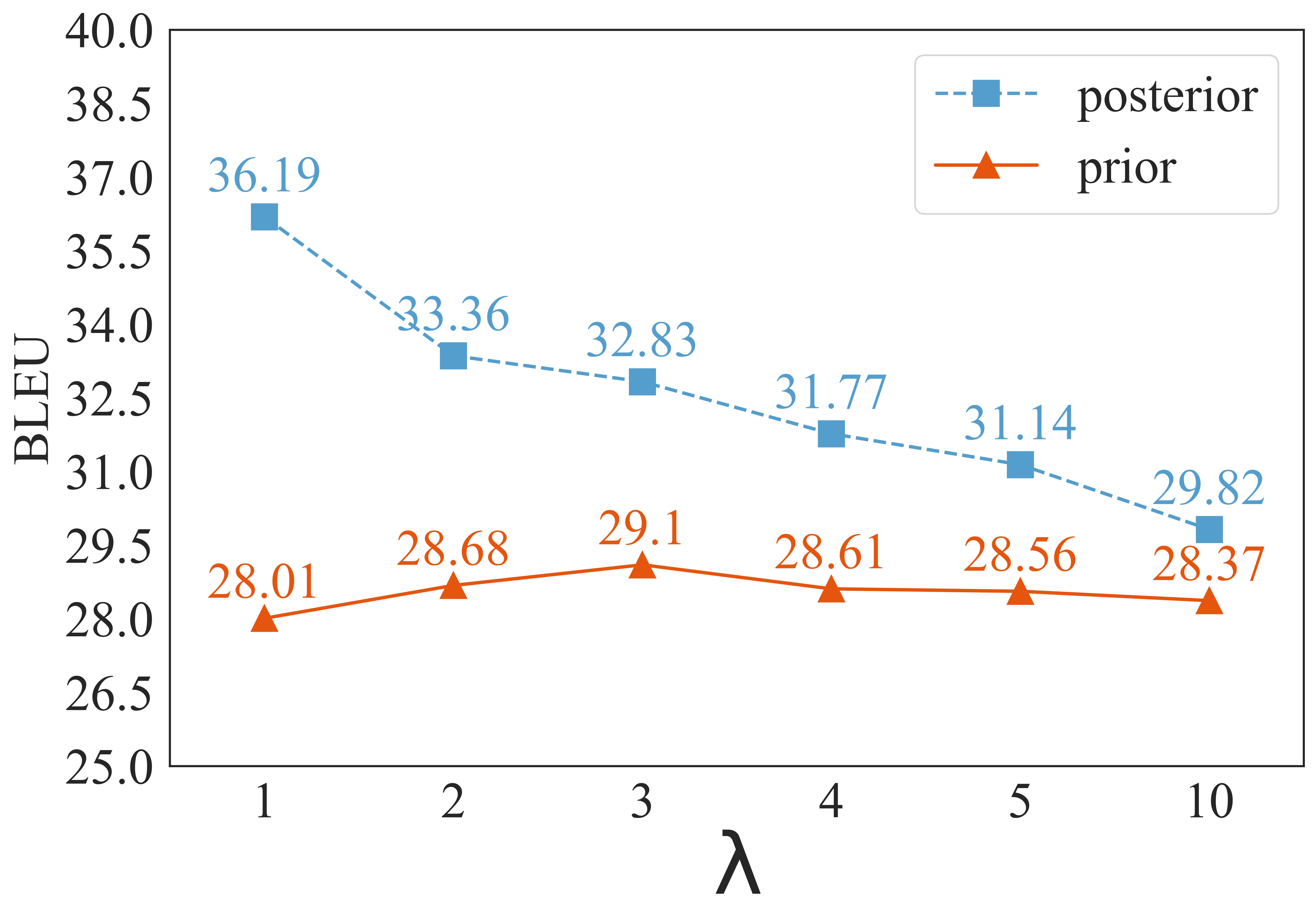} 
\caption{BLEU scores curves for prior and posterior paths against $\lambda$ of $L_{kl}$ on Dev set of PHOENIX14T.}
\label{fig_exp_ablation_BLEU_Cruve}
\end{figure}

Predicting the target variable $y$ is much easier for the posterior path, as it is also provided as input to the encoder. Therefore, we argue that the posterior serves as an upper bound for the prior, given that ground truth textual information is available. 
We anticipate that incorporating a self-distillation term ($\mathcal{L}_{\text{SD}}$) in our loss function will encourage prior to learn from the posterior; however, this may compromise the performance of the posterior at the same time. 
Thus, a trade-off exists between optimizing both prior and posterior. As illustrated in Fig. \ref{fig_exp_ablation_BLEU_Cruve}, as the weight parameter $\lambda$ increases, the performance of the posterior model consistently deteriorates, while the performance of the prior model initially reaches a peak and subsequently declines.


\begin{figure}[tb]
\centering
\includegraphics[width=0.4 \textwidth]{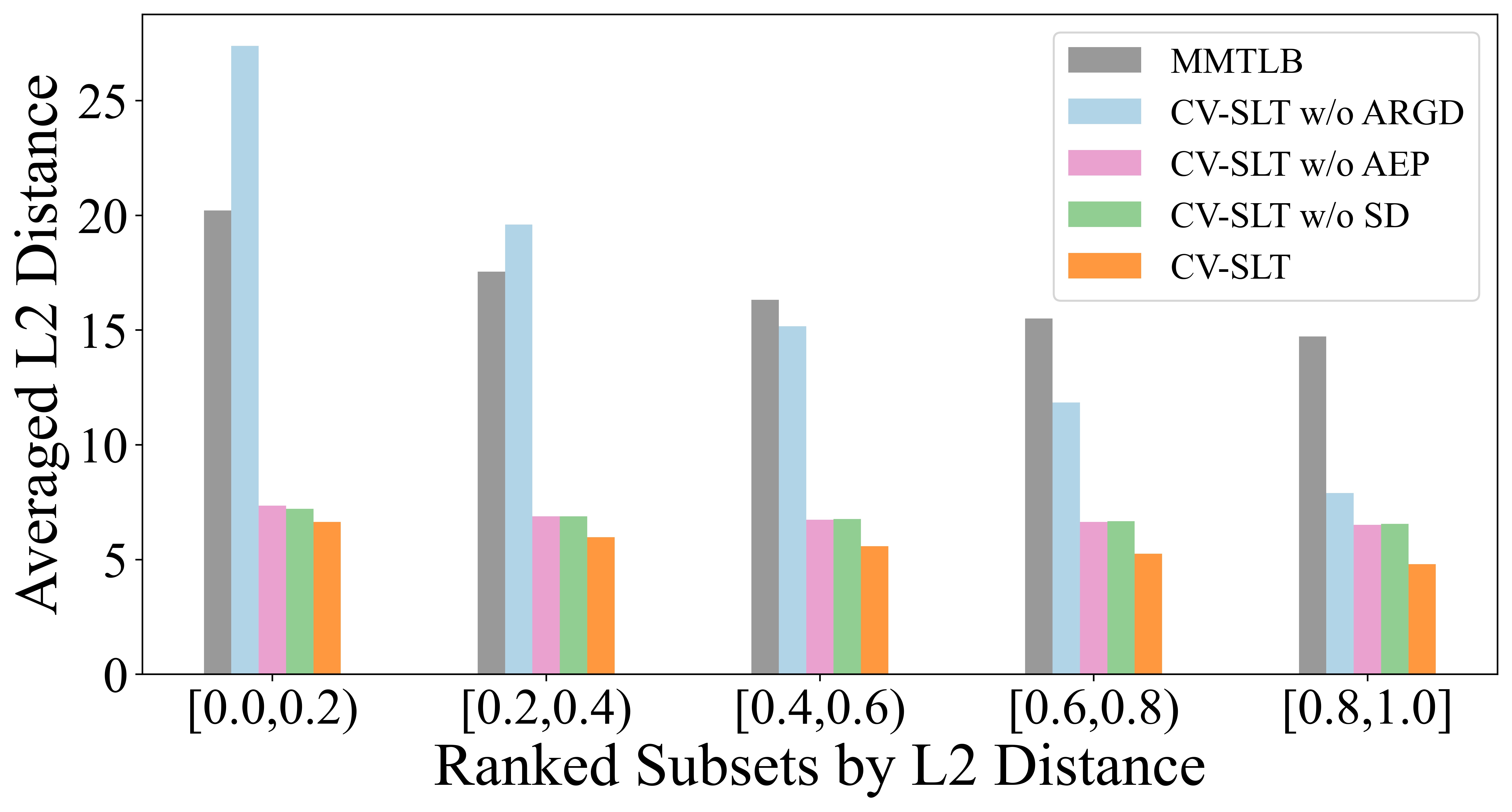} 
\caption{Visualization on effectiveness of modal alignment for each component.}
\label{ablation_modality_alignment}
\end{figure}

\subsubsection{Is the modality gap mitigated?}

\begin{figure}[ht]
\centering
\includegraphics[width=0.4 \textwidth]{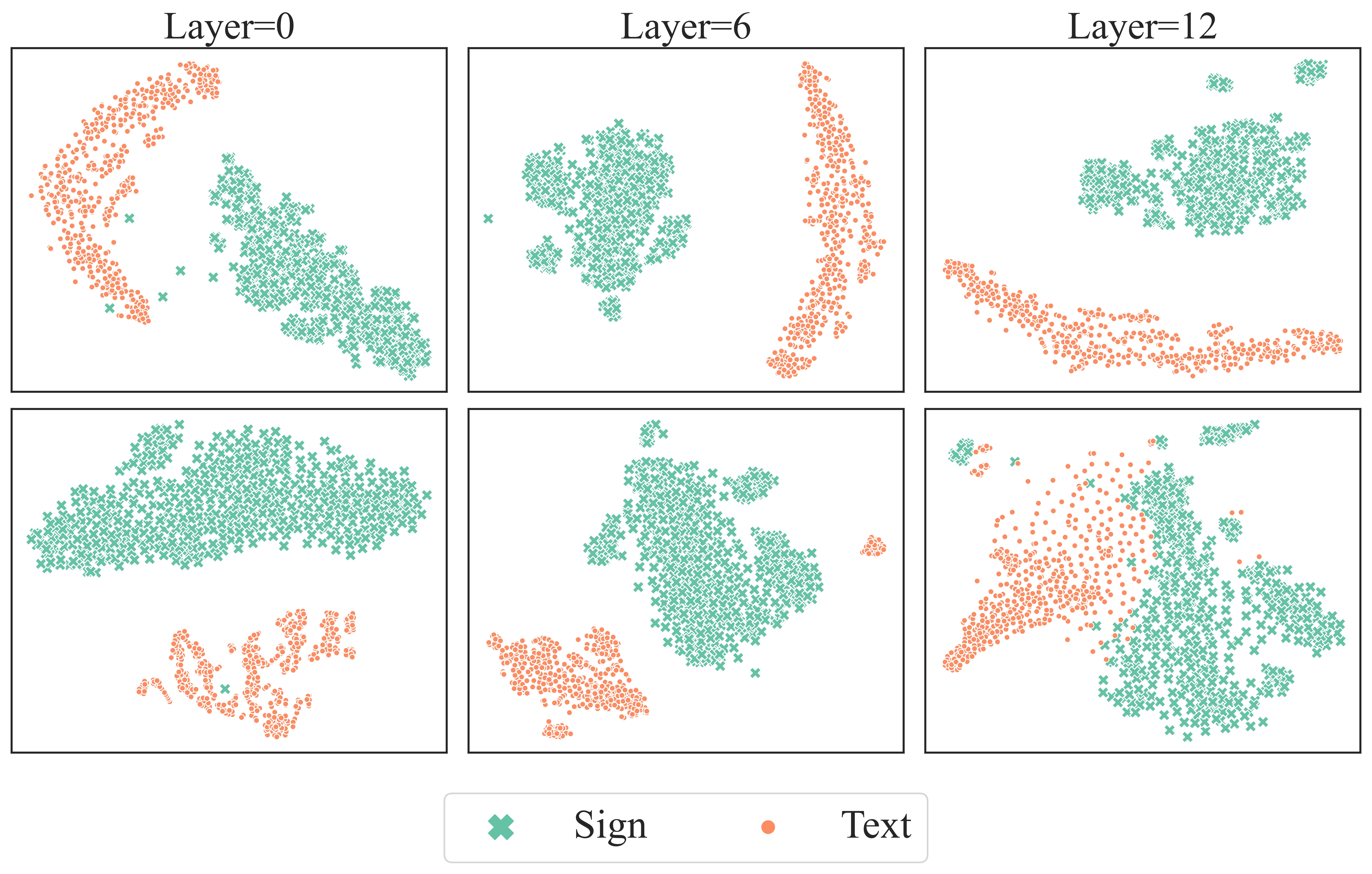} 
\caption{2-D illustration of the sentence-level representations of different layers of the encoder outputs. Upper: MMTLB; Bottom: Our CV-SLT.}
\label{ablation_aligned_tsne}
\end{figure}

Our CV-SLT aims to mitigate the representation discrepancy and bridge the modality gap between sign language and target text, as SLT is a typical cross-modal task. 
To verify the cross-modal alignment ability of our CV-SLT, we first demonstrate the effectiveness of modal alignment for each component in Fig \ref{ablation_modality_alignment}. Consistent with the preliminary experimental setup, L2 distance is used as a statistical metric for the modality gap. The proposed ARGD module brings the most contribution while other components are indispensable as well. 

We additionally draw the 2-D data distribution of sign representation and text representation with T-SNE \cite{maatenVisualizingDataUsing2008} to show the alignment comprehensively.
Fig. \ref{ablation_aligned_tsne} plots the sentence-level encoder output of two modalities on Dev set of PHOENIX14T. MMTLB (upper bound) demonstrates a clear division between the two representations, indicating that it independently processes these two types of inputs with a faintish alignment.
Conversely, CV-SLT (lower bound) exhibits a different behaviour where the representations of both modalities gradually converge towards each other with increasing encoder layers.

\section{Related works}

\subsection{Sign Language Translation}

\citeauthor{camgozNeuralSignLanguage2018} \shortcite{camgozNeuralSignLanguage2018} first propose the PHOENIX14T dataset and start to explore directly translating from sign language videos to spoken language text. However, unlike NMT where both the source and target texts are in written form, SLT involves sign language videos as the input and necessitates translation into textual sentences. This inherently presents a challenge due to the modality gap between sign language videos and target text. Therefore, a variety of methods were proposed to tackle this problem.
\citeauthor{cihancamgozSignLanguageTransformers2020} \shortcite{cihancamgozSignLanguageTransformers2020} proposed to jointly train SLR and SLT with the gloss annotation serving as supervision to regularize the outputs transformer encoder. \citeauthor{zhouImprovingSignLanguage2021} \shortcite{zhouImprovingSignLanguage2021} use a back-translation strategy to generate pseudo text-gloss-sign pairs for data augmentation. 
\citeauthor{chenSimpleMultiModalityTransfer2022} \shortcite{chenSimpleMultiModalityTransfer2022} leverage extensive external knowledge, such as human action and spoken languages, through a progressive pretraining approach that involves Sign2Gloss and Gloss2Text modules to be trained from general domains to within sign language domains. 

Another mainstream is to use complex multi-cues including information from hand shapes, facial expressions, mouths, and poses to enhance the performance of visual module \cite{camgozMultichannelTransformersMultiarticulatory2020,zhengEnhancingNeuralSign2021,zhouSpatialTemporalMultiCueNetwork2022,chenTwoStreamNetworkSign2022}.

\subsection{Variational Alignment}
\citeauthor{kingmaAutoEncodingVariationalBayes2014b} \shortcite{kingmaAutoEncodingVariationalBayes2014b} and 
\citeauthor{rezendeStochasticBackpropagationApproximate2014} \shortcite{rezendeStochasticBackpropagationApproximate2014}
first introduce variational autoencoders (VAE).Typically, these models introduce a neural inference model to approximate the intractable posterior, and optimize model parameters jointly with Stochastic Gradient Variational Bayes. Conditional variational auto-encoder (CVAE) \cite{sohnLearningStructuredOutput2015} is a modification of VAE to generate text or image conditioned certain given attributes. VAE has achieved great success in the NMT community, 
\citeauthor{zhangVariationalNeuralMachine2016} \shortcite{zhangVariationalNeuralMachine2016} and 
\citeauthor{suVariationalRecurrentNeural2018} \shortcite{suVariationalRecurrentNeural2018} 
introduce latent variables to NMT to complement the undesirable learned attentions and learn the shared semantic alignment between bilingual sentence pair. 
\citeauthor{shuLatentVariableNonAutoregressiveNeural2020} \shortcite{shuLatentVariableNonAutoregressiveNeural2020}, \citeauthor{zhuNonAutoregressiveNeuralMachine2022} \shortcite{zhuNonAutoregressiveNeuralMachine2022} and \citeauthor{baoLatentGLATGlancingLatent2022} \shortcite{baoLatentGLATGlancingLatent2022} 
propose to refine the latent variables rather than target tokens to mitigate the multi-mode problem of non-autoregressive translation task (NAT), resulting better BLEU score. 

In the sign language-related domain, \citeauthor{zhengCVTSLRContrastiveVisualTextual2023} \shortcite{zhengCVTSLRContrastiveVisualTextual2023} first introduce VAE to sign language recognition (SLR) task to help align the sign language videos and gloss representation. They first train a variational auto-encoder to capture the gloss-related distribution and then cascade the off-the-shelf visual module and textual module with a video-gloss adapter to perform the SLR task. Differently, we are concentrated on aligning the sign representation videos and spoken language text rather than the intermediate gloss representation, in a more direct way rather than a multi-stage strategy.

\section{Conclusion}
In this paper, we propose CV-SLT, a  sign language translation framework based on conditional variational autoencoder, which aims to bridge the cross-modal representation discrepancy between sign language and spoken language text. 
To achieve this, we introduce prior and posterior paths to model the marginal distribution of visual modality and the joint distribution of both visual and textual modalities. 
Two Kullback-Leibler divergences are utilized to regularize the encoder outputs and decoder outputs, ensuring the consistency of prior and posterior.
In the future, we are interested in introducing discrete latent variables instead of continuous ones for better intermediate representation for SLT.

\section{Acknowledgments}
We thank all the anonymous reviewers for their insightful and valuable comments. This work was supported by the National Natural Science Foundation of China ``Research on Neural Chinese Sign Language Translation Methods Integrating Sign Language Linguistic Knowledge'' (No. 62076211) and Central Leading Local Project ``Fujian Mental Health Human-Computer Interaction Technology Research Center'' (No. 2020L3024).

\bibliography{aaai24}

\end{document}